\documentclass[10pt,twocolumn,letterpaper]{article}

\usepackage{cvpr}
\usepackage{times}
\usepackage{epsfig}
\usepackage{graphicx}
\usepackage{amsmath}
\usepackage{amssymb}
\usepackage{comment}
\usepackage{times}
\usepackage{epsfig}
\usepackage{graphicx,dblfloatfix}
\usepackage{mwe}
\usepackage{subfig}
\usepackage{amsmath}
\usepackage{amssymb}
\usepackage{authblk}
\usepackage{url}

\usepackage{authblk}
\usepackage[ruled,norelsize]{algorithm2e} 
\usepackage{breakurl}
\usepackage{array,caption,tabularx,makecell,booktabs} 

\usepackage[font=footnotesize, labelfont=footnotesize, labelfont=bf]{caption}

\usepackage[pagebackref=false,breaklinks=true,letterpaper=true,colorlinks,bookmarks=false]{hyperref}

\usepackage[breaklinks=true,bookmarks=false]{hyperref}

\cvprfinalcopy %

\def\cvprPaperID{****} %

\ifcvprfinal\fi
\makeatletter
\renewcommand\AB@affilsepx{, \protect\Affilfont}
\makeatother
\begin{document}

\title{IsMo-GAN: Adversarial Learning for Monocular Non-Rigid 3D Reconstruction \vspace{-15pt}} 
\author{ 
  \hspace{-0pt} Soshi Shimada$^{1, 2}$ $\;\;\;\;\;\;$ Vladislav Golyanik$^{1, 3}$ $\;\;\;\;\;\;$ Christian Theobalt$^{3}$ $\;\;\;\;\;\;$ Didier Stricker$^{1, 2}$ \vspace{1pt}\\
  \hspace{-35pt}$^{1}$University of Kaiserslautern $\;\;\;\;\;\;\;\;\;\;\;$ $^{2}$DFKI $\;\;\;\;\;\;\;\;\;\;\;$ $^{3}$MPI for Informatics 
} 

\maketitle
\begin{abstract} 
The majority of the existing methods for non-rigid 3D surface regression from monocular 2D images require an object template or point tracks over multiple frames as an input, and are still far from real-time processing rates. In this work, we present the %
Isometry-Aware Monocular Generative Adversarial Network (IsMo-GAN) --- %
an approach for direct 3D reconstruction from a single image, trained for the deformation model in an adversarial manner on a light-weight synthetic dataset. 
IsMo-GAN reconstructs surfaces from real images under varying illumination, camera poses, textures and shading %
at over $250$ \textit{Hz}. 
In multiple experiments, it consistently outperforms several approaches in the reconstruction accuracy, runtime, generalisation to unknown surfaces and robustness to %
occlusions. %
In comparison to the state-of-the-art, we reduce the reconstruction error by $10$-$30\%$ including the textureless case and our surfaces evince fewer artefacts qualitatively. 
\end{abstract} 
\vspace{7pt}

\section{Introduction} 

\begin{figure}[t]
 \centering
  \includegraphics[width=1.0\linewidth]{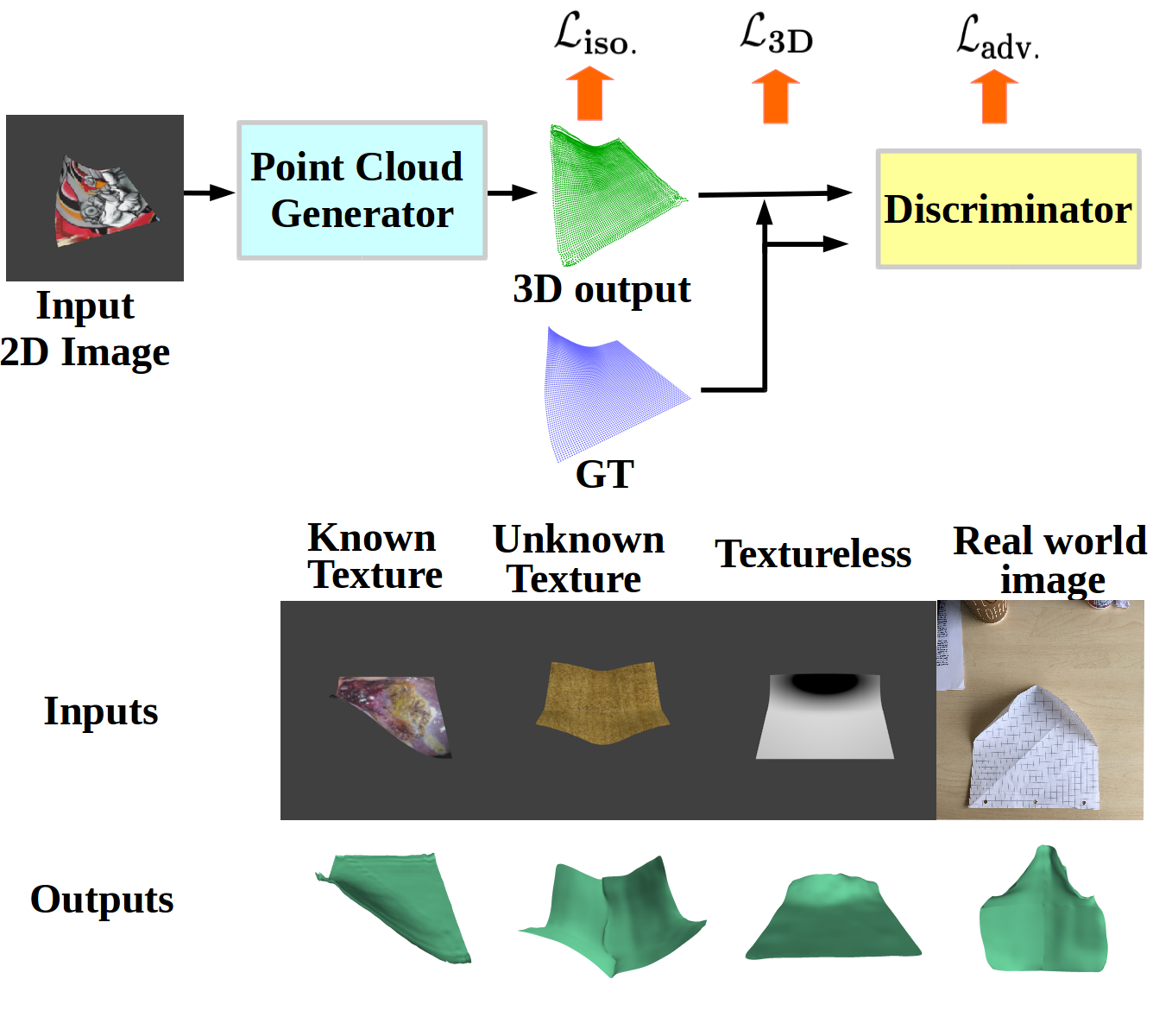} 
  \caption{Overview of our IsMo-GAN approach. (top) The \textit{generator} network accepts a 2D RGB image segmented by the \textit{object detection} network (OD-Net) and returns a 3D point cloud. The output and ground truth (GT) are fed to the \textit{discriminator} network which serves as a surface regulariser. 
  (bottom) Example reconstructions by IsMo-GAN in different scenarios: a known texture, an unknown texture, a textureless surface and a reconstruction of a real image.} \label{fig:simpleoverview} 
  \vspace{8pt}
\end{figure}

Monocular non-rigid 3D reconstruction from single 2D images is a challenging ill-posed problem in computer vision with many useful applications. 
Such factors as varying illumination, external and self occlusions in the scene and lack of texture further complicate the setting. %
In recent times, dense monocular non-rigid reconstruction was mostly tackled by shape-from-template (SfT) techniques and non-rigid structure from motion (NRSfM). SfT requires a template --- an accurate geometry estimate corresponding to one of the 2D views known in advance \cite{Salzmann2008, Perriollat2010, Bartoli2012, Oestlund2012, haouchine2014, Yu2015} ---, whereas NRSfM relies on motion and deformation cues in the input point tracks over multiple views \cite{Bregler2000, Torresani2008, Gotardo2011, Garg2013, Paladini2012, Golyanik2017d, Kumar2018}. Currently, there is a lack of approaches supporting real-time processing rates which is a desired property for interactive applications. %

At the same time, convolutional neural networks (CNN) \cite{Lecun98} have been successfully applied in various domains of computer vision including, among other architectures, fully convolutional encoder-decoders to convert data modalities, as in object segmentation and contour detection  \cite{badrinarayanan2017segnet, chen2016combining, chen2017convolutional, Hongsheng2014}. 
Many applications benefit from the properties of different modifications of generative adversarial networks (GAN) %
\cite{isola2017image,karras2017progressive,mirza2014conditional,radford2015unsupervised,shrivastava2017learning,zhu2017unpaired}. 
GAN include two competing neural networks which are trained simultaneously during the training phase --- the \textit{generator} and \textit{discriminator} networks. %
Starting from arbitrary signals, the generator mimics data distributions of the training dataset and %
learns to pass the discriminator's test on sample authenticity. %
The discriminator estimates the probabilities that given outputs originate from the training dataset or from the generator. 
This adversarial manner allows the generator to pursue a high-level objective, \textit{i.e,} \textit{``generate outputs that look authentic and have the properties of the representative samples"}. 

In this paper, we propose \textit{Isometry-Aware Monocular Generative Adversarial Network} (IsMo-GAN) --- a framework with several CNNs for the recovery of a deformable 3D structure from 2D images, see Fig.~\ref{fig:simpleoverview} for an overview. Our approach learns a deformation model, and the individual CNNs are trained in an adversarial manner to enable generalisation to unknown data and robustness to noise. %
In the 3D reconstruction task, the adversarial training is targeted at the objective \textit{``generate realistic 3D geometry"}. This high-level objective improves the reconstruction qualitatively because lower Euclidean distances between the predicted and ground truth geometry do not necessarily imply higher visual quality.

\subsection{Contributions} 

By combining a CNN with skipping connections for 3D reconstruction, an adversarial learning (a discriminator and geometry regulariser) and a confidence map indicator for object segmentation, we develop an approach that directly regresses 3D point clouds while consistently outperforming competing methods \cite{Garg2013, Yu2015, Ngo_2015_ICCV, liu2017better, Golyanik2017d, golyanik2018hdm, bednarik2018learning} quantitatively by $10$-$30\%$ across various experiments and scenarios (see Fig.~\ref{fig:architecture} and Sec.~\ref{sec:Experiment}). 
IsMo-GAN enhances the \textbf{reconstruction accuracy of real images} compared to the competing methods, 
including the regression of \textbf{textureless surfaces}. 
The demonstrated improvement is due to the key technical \textbf{contributions} of the method --- first, \textbf{the adversarial regulariser loss} and, second, the integrated \textbf{object detection network} (OD-Net) for the foreground-background segmentation, %
as we show in the comparison with the most closely related previous method \cite{golyanik2018hdm} (refer to Sec.~\ref{sec:Experiment}). 

IsMo-GAN does not require a template, camera calibration parameters or point tracks over multiple frames. %
Our pipeline is robust to varying illumination and camera poses, internal and external occlusions and unknown textures, and all that with a training on light-weight datasets of non-rigid surfaces \cite{golyanik2018hdm, bednarik2018learning}. 
Concerning the runtime, IsMo-GAN exceeds conventional methods by a large margin and reconstructs \textbf{up to 250 states per second}. 
Compared to computationally expensive 3D \cite{Maturana2015, choy2016, Riegler2017} and graph convolutions \cite{Defferrard2016, Verma2018}, IsMo-GAN applies 2D convolutions \cite{Krizhevsky2012} for 3D surface regression from 2D images. %
To the best of our knowledge, our study is the first one for deformation model-aware non-rigid 3D surface regression from single monocular images with point set representation trained in an adversarial manner and a masking network in a single pipeline. %

\subsection{Paper Structure} 

The rest of the paper is organised as follows. In Sec.~\ref{sec:related_work}, we discuss related works. Technical details and the network architectures are elaborated in Sec.~\ref{sec:Method}. Sec.~\ref{sec:Experiment} describes the experiments. 
Finally, we discuss the method including its limitations in Sec.~\ref{sec:Limitations} and summarise the study in Sec.~\ref{sec:Conclusion}. 

\section{Related Work}\label{sec:related_work}

In this section, we review the most related model-based (Sec.~\ref{ssec:conventional_methods}) and deep neural network (DNN)-based techniques  (Secs.~\ref{ssec:DNN_based}--\ref{ssec:AdvTrain}). 

\subsection{Unsupervised Learning Methods}\label{ssec:conventional_methods} %

NRSfM factorises point tracks over multiple views into camera poses and non-rigid shapes relying on motion and 
deformation cues as well as weak prior assumptions (\textit{e.g.,} temporal state smoothness or expected deformation complexity) 
\cite{Bregler2000, Torresani2008, Gotardo2011, Garg2013, Kumar2018}. Only recently NRSfM has entered the realm of dense reconstructions \cite{Russell2012, Garg2013, Agudo2014, Golyanik2017d}. %
Dense NRSfM requires distinctive textures on the target object during the tracking phase \cite{Garg2010, Taetz2016, Li2016}. 
Even though the reconstruction can be performed at interactive rates \cite{Agudo2014}, obtaining dense correspondences from real images can significantly decrease the overall throughput of the pipeline. 
The recent work of Gallardo \textit{et al.}~\cite{gallardo2017dense} can cope with textureless objects by considering shading and still, their solution 
is computationally expensive. 
IsMo-GAN reconstructs textureless objects upon the learned deformation model while fulfilling the real-time requirement. %

SfT, also known as non-rigid 3D tracking, requires a 3D template known in advance, \textit{i.e.,} an accurate reconstruction with given 2D-3D correspondences \cite{Salzmann2008, Bartoli2012, Yu2015}. %
Several approaches enhance robustness of SfT to illumination changes with the shape-from-shading component \cite{Malti2012, liu2017better}. 
Our method does not require a template --- all we need as an input is a single monocular 2D image during the surface inference phase. 
At the same time, IsMo-GAN is trained in the supervised manner. The training dataset contains a %
sequence of 3D states along with the corresponding 2D images \cite{golyanik2018hdm}. %
Thus, our framework bears a remote analogy with SfT, as IsMo-GAN is trained for a deformation model %
with a pre-defined surface at rest (or multiple surfaces at rest, in the extended version). 

\subsection{DNN-Based 3D Reconstruction Techniques}\label{ssec:DNN_based}

Methods for 3D reconstruction with DNNs primarily focus on rigid scenes \cite{wu2016learning,jiang2018gal,gwak2017weakly, Eigen_2014, choy2016, Godard2017, Riegler2017, Kurenkov2018} while only a few approaches were proposed for the non-rigid case so far \cite{golyanik2018hdm, pumarola2018geometry}. 
Volumetric representation is often used in DNN based approaches \cite{Maturana2015, choy2016, Riegler2017}. %
In most cases, it relies on computationally costly 3D convolutions limiting the techniques in the supported resolution and throughput. %
Qualitatively, volumetric representations lead to  discretisation artefacts. 
Our approach directly regresses 3D point coordinates by applying computationally less expensive 2D convolutions \cite{Lecun98, Krizhevsky2012}, 
and surfaces recovered by \hbox{IsMo-GAN} are smoother and more realistic qualitatively. %
 Golyanik \textit{et al.}~\cite{golyanik2018hdm} recently proposed Hybrid Deformation Model Network (HDM-Net) for monocular non-rigid 3D reconstruction targeting virtual reality applications. In their method, an encoder-decoder network is trained for a deformation model with a light-weight synthetic dataset of thin plate states in the point cloud representation. %
 Rather than treating every image as a different rigid instance of a pre-defined object class \cite{Kanazawa2018}, HDM-Net associates every input image with a non-rigid surface state 
 imposing the isometry and feasibility constraint upon the learned deformation model. In addition, its objective function includes a contour loss. We do not use the contour loss as it increases the training time and does not make a significant difference in the reconstruction accuracy. %
 We regress $50$ states per second more on average with a higher accuracy compared to HDM-Net \cite{golyanik2018hdm}. %
 Moreover, IsMo-GAN shows more accurate results for occluded and textureless surfaces as well as when reconstructing from real images. 

 Pumarola \textit{et al.}~\cite{pumarola2018geometry} %
 combine three sub-networks for 2D heat-map generation with object detection, depth estimation and 3D surface regression. %
 For the real-world scenario, they have to finetune the pipeline. %
 In contrast, IsMo-GAN automatically segments and reconstructs real images, with no need for further parameter tuning. 
 Bedna\v{r}\'{i}k \textit{et al.}~\cite{bednarik2018learning} employ a trident network with a single encoder and three decoders for %
 the depth-map, normal map and 3D mesh estimation. For mesh decoding, they use a fully-connected layer. 
 Similar to \cite{golyanik2018hdm, pumarola2018geometry}, our generator consists of 2D convolutional layers and includes multiple sub-networks. 
 In contrast, IsMo-GAN uses an adversarial loss which leads to the consistently improved accuracy across different scenarios. 
 \begin{figure*}[t!]
 \centering
  \includegraphics[width=0.97\linewidth]{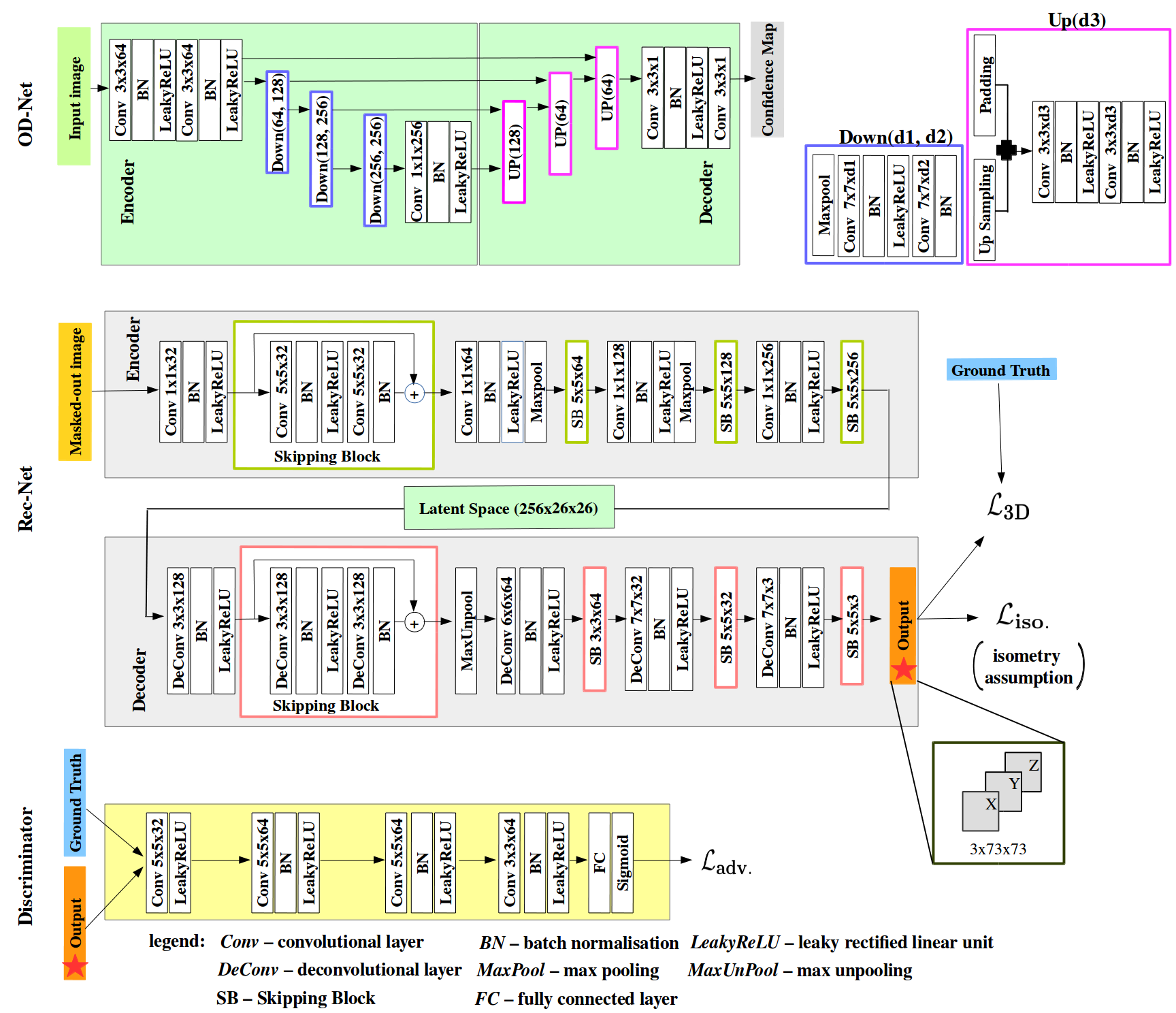} 
  \caption{ Architecture of the proposed IsMo-GAN framework.     %
  \textit{Up Sampling} in OD-Net doubles the width and height of the input using binary interpolation. OD-Net applies \textit{padding} on the inputs to equalise the input dimensionalities if necessary. Rec-Net accepts images of the size $224 \times 224 \times 3$ (with three colour channels). The output is a $73 \times 73 \times 3$ dense reconstruction, with $73^2$ points per frame. The fully-connected layer in the discriminator converts the dimensionality from $3136$ to $1$ in order to generate the probabilistic decision about the input authenticity (the activation from the fourth convolutional layer is of the dimension $7 \times 7 \times 64$ leading to the dimensionality $3136$ when concatenated). } \label{fig:architecture} 
\end{figure*}

\subsection{Adversarial Learning in Computer Vision}\label{ssec:AdvTrain} %
GAN were initially introduced as a generative model for the sampling of new instances from a predefined class \cite{goodfellow2014generative}.
In GAN, learning to sample from a training distribution is performed through a two-player game and formalised as the adversarial loss. %
GAN were applied for various tasks including image inpainting \cite{pathakCVPR16context,yu2018generative}, video generation \cite{wu2017gp,vondrick2016generating}, 2D image resolution enhancement \cite{ledig2017photo,wang2018esrgan}, image texture transfer \cite{li2016precomputed} and a transfer from texts to images \cite{zhang2017stackgan}, among others. %
Several improvements for training convergence and performance of GAN were subsequently proposed over the last years \cite{isola2017image,mirza2014conditional,radford2015unsupervised,zhu2017unpaired}.
The adversarial loss is also applicable as a fidelity regulariser in rigid 3D reconstruction \cite{jiang2018gal}. %
In \cite{jiang2018gal}, the conditional adversarial loss demands the inference result to be close to the shape probability distribution of the training set.
Adversarial loss in IsMo-GAN targets the deformation model of a thin structure instead of the space of multiple shapes, \textit{i.e.,} the recovered surfaces are constrained to be reasonable with respect to the probability distribution of the learned space of non-rigid states. To the best of our knowledge, it is the first time an adversarial loss is applied in monocular non-rigid surface reconstruction with DNNs. %
 \section{The Proposed Method}\label{sec:Method} 

In this section, we first describe the proposed architecture (Sec.~\ref{ssec:architecture}) followed by the loss functions  (Sec.~\ref{ssec:loss_functions}). Next, we provide details about the dataset (Sec.~\ref{ssec:data_set}) and IsMo-GAN training (Sec.~\ref{ssec:training_details}). %

\subsection{Network Architecture}\label{ssec:architecture} 

We propose a DNN architecture that consists of a generator and discriminator networks, see Fig.~\ref{fig:architecture} for the schematic visualisation. The generator is, in turn, composed of OD-Net and Reconstruction Network (Rec-Net), both based on an encoder-decoder architecture with skipping connections \cite{He2016}. 
The input images are of the resolution $224 \times 224$. 
OD-Net has a U-net structure \cite{ronneberger2015u, CusUNet}, and it is responsible for the generation of a grayscale confidence map indicating the position of the target object. 
The generated confidence map is subsequently binarised \cite{4310076} and the target object is extracted with the algorithm of Suzuki \textit{et al.}~\cite{suzuki1985topological}. 
Compared to the customised U-Net \cite{CusUNet}, the number of downsampling and upsampling convolutional blocks is reduced by one in our OD-Net due to the relatively small size of the training dataset (see Sec.~\ref{ssec:data_set}). Rec-Net is a residual encoder-decoder network. The encoder extracts relevant features for 3D reconstruction from the given 2D inputs and converts them into the latent space 
representation. The decoder increases the dimensionality of the latent space in height and width and adjusts the depth of the latent space until its activation reaches the dimensionality of $73 \times 73 \times 3$, \textit{i.e.,} the dimensionality of the ground truth training states. 

Our discriminator consists of four blocks --- a convolutional layer, leaky rectified linear unit (ReLU) \cite{maas2013rectifier}, batch normalisation and a fully-connected layer. To enhance training stability, the first layer set of the discriminator does not contain batch normalisation \cite{radford2015unsupervised}. The output from Rec-Net is evaluated by several loss functions. 
First, we penalise Euclidean distances between the ground truth 3D geometry and output of the generator with the sum of absolute differences (SAD). 
Next, similar to \cite{golyanik2018hdm}, we assume the observed surfaces to be isometric and introduce a soft isometry constraint, \textit{i.e.,} a loss function penalising the roughness and non-isometric effects (\textit{e.g.,} shrinking and dilatation) of the predicted 3D geometry in an unsupervised manner. 
For more plausible and realistic outputs, we introduce an adversarial loss \cite{goodfellow2014generative} which targets the deformation model of a %
surface. In the following section, all three losses of IsMo-GAN are described in detail. %

\subsection{Loss Functions}\label{ssec:loss_functions}

Suppose $\mathbf{I} = \{\mathbf{I}_m^{n}\}$, $m \in \{1, \hdots, M\}$, $n \in \{1, \hdots, N\}$ denote 2D input images, with 
the total number of states $M$ and the total number of images for each state $N$. 
Let $\mathbf{S}^{\text{GT}} = \{\mathbf{S}_{m}^{\text{GT}}\}$ be the ground truth geometry. $\bf{G}$ and $\bf{D}$ denote the generator (Rec-Net) and  discriminator components. The total loss of IsMo-GAN reads: 
\begin{equation}
\small
\hspace{-5pt}
\begin{aligned}
   \mathbf{\mathcal{L}}(\mathbf{I}, \mathbf{S}^{\text{GT}}) = \mathbf{\mathcal{L}_{adv.}}(\mathbf{I},\mathbf{S}^{\text{GT}}) + \mathbf{\mathcal{L}_{iso.}}(\mathbf{G(I)}) 
+ \mathbf{\mathcal{L}_{3D}}(\mathbf{G(I)}, \mathbf{S}^{\text{GT}}),
\end{aligned}
\end{equation}
where $\mathbf{G(I)}$ stands for the reconstructed 3D surfaces.

\paragraph{3D Loss.} 
The 3D loss is based on SAD function which penalises the Euclidean distance between ground truth geometry and the predicted 3D geometry per point: 
\begin{equation}\label{eq:loss_3D}
\mathbf{\mathcal{L}_{3D}}(\mathbf{G(I)}, \mathbf{S}^{\text{GT}})=\frac{1}{MN}\sum_{m=1}^{M}\sum_{n=1}^{N}\lvert\mathbf{S}_{m}^{\text{GT}} - \mathbf{G(I}_m^{n})\rvert.
\end{equation}
\paragraph{Isometry Prior.} 
The isometry prior penalises surface roughness. We assume the target object to be isometric which implies that every 3D point has to be located close to the neighbouring points. This loss was already effectively applied in HDM-Net \cite{golyanik2018hdm}. The corresponding  loss function is expressed in terms of the difference between the predicted geometry and its smoothed version: %
\begin{equation}\label{eq:isometry_loss} 
\mathbf{\mathcal{L}_{iso.}(G(I))} = \frac{1}{MN}\sum_{m=1}^{M}\sum_{n=1}^{N} \lvert \hat{\mathbf{S}}_m^{n} - \mathbf{G(I}_m^{n}) \rvert. 
\end{equation}
In Eq.~\eqref{eq:isometry_loss}, $\hat{\mathbf{S}}_m^{n}$ denotes the surface smoothed by a Gaussian kernel: 
\begin{equation}
\hat{\mathbf{S}}_m^{n}=\frac{1}{2\pi\sigma^{2}} \, \operatorname{exp}\left ( -\frac{x^{2}+y^{2}}{2\sigma^{2}}\right)\ast\mathbf{G(I}_m^{n}), 
\end{equation}
where $\ast$ is the convolution operator, $\sigma$ is the standard deviation of the Gaussian kernel, and 
$x$ and $y$ stand for the point coordinates.

\paragraph{Adversarial Loss.} 
As an objective function of the adversarial training, we employ binary cross entropy (BCE) \cite{goodfellow2014generative} defined as 

\begin{equation}\label{eq:generator_loss}
\begin{aligned}
\mathbf{\mathcal{L}_{G}(I)}=-\frac{1}{MN}\sum_{m=1}^{M}\sum_{n=1}^{N}\,\log(\mathbf{D}(\mathbf{G}(\mathbf{I}_m^{n}))  
\end{aligned}
\end{equation}
for the generator, and 
\begin{equation}\label{eq:discriminator_loss}
\footnotesize
\mathbf{\mathcal{L}_{D}}(\mathbf{I},\mathbf{S}^{\text{GT}})=-\frac{1}{MN}\sum_{m=1}^{M}\sum_{n=1}^{N} \, \big[\log(\mathbf{D}(\mathbf{S}_{m}^{\text{GT}}))+\log(1-\mathbf{D(G(I}_m^{n}))\big] 
\end{equation}
for the discriminator. 
The adversarial loss is then comprised of the sum of both components: 
\begin{equation}\label{eq:generative_loss}
\mathbf{\mathcal{L}_{adv.}}(\mathbf{I},\mathbf{S}^{\text{GT}})=\mathbf{\mathcal{L}_{G}}(\mathbf{I}) + \mathbf{\mathcal{L}_{D}}(\mathbf{I},\mathbf{S}^{\text{GT}}).
\end{equation}
The adversarial loss in Eq.~\eqref{eq:generative_loss} defines the high-level goal that encourages IsMo-GAN to generate visually more realistic surfaces.  %
It is the core component which enables IsMo-GAN to outperform HDM-Net \cite{golyanik2018hdm} by $10-15\%$ quantitatively as well as qualitatively on real images (see Sec.~\ref{ssec:synth_data_experiments}). 

We observed that using SAD as 3D loss tends to propagate the surface roughness from the input to the output. The isometry prior reduces the roughness, slightly shrinks the output and smoothes the corners. The adversarial loss compensates for these undesired effects of the 3D loss and the isometry prior, and serves as a novel regulariser for surface deformations. %

\subsection{Training Datasets}\label{ssec:data_set} 

In this section, we elaborate on the main datasets \cite{golyanik2018hdm, kolesnikov2014closed} used to train the OD-Net, Rec-Net and the discriminator. 
In Sec.~\ref{subs:textureless}, we extra use the \textit{textureless cloth} dataset \cite{bednarik2018learning} to train a variation of our pipeline and compare its performance on textureless surfaces.

\subsubsection{Deformation Model Dataset}\label{sssec:deformation_model_dataset} 

We use the synthetic 2D-3D thin plate dataset from \cite{golyanik2018hdm} for the training and tests. 
In total, the dataset contains $4648$ states representing different isometric non-linear deformations of a thin plate structure 
(\textit{e.g.,} waving deformations and bending). %
Due to the original ${4}{:}{1}$ training-test split, $M = 3728$, and $N = 60$ 
(three textures illuminated by a light source at four different locations, and each combination of the texture and illumination is rendered with five virtual cameras). 
Every 3D state contains $73^2$ 3D points sampled on a regular grid at rest, with a consistent topology across all states. 
For each 3D state, there are corresponding rendered 2D images of the resolution $256 \times 256$\footnote{the input images are resized to $224 \times 224$ in our pipeline} for the combinations with five different positions of the light source, four different textures (\textit{endoscopy}, \textit{graffiti}, \textit{clothes} and \textit{carpet}) and five different camera poses. 
To train IsMo-GAN and competing methods for the shape-from-shading, we extend the thin plate dataset \cite{golyanik2018hdm} with a subsequence 
of deforming textureless surfaces (the states are left the same while the texture is removed). %
In our dataset extension, $M = 3728$ and $N = 5$ (no texture, five virtual cameras).

\subsubsection{OD-Net Dataset}\label{sssec:od_net_dataset} 

To train OD-Net, we generate a mixed image dataset with varying backgrounds (\textit{sky}, \textit{office}  and \textit{forest}) and the corresponding binary masks. First, we randomly translate the target object in the images from the deformation model dataset (Sec.~\ref{sssec:deformation_model_dataset}). 
Next, we combine the first part with a dataset of real-world RGB images and the corresponding binary masks from  \cite{kolesnikov2014closed}. In total, our mixed dataset contains $\approx 14k$ images and corresponding binary masks.

\subsection{Training Details}\label{ssec:training_details} 

We use Adam \cite{kingma2014adam} for optimisation of network parameters, with the learning rate of $10^{-3}$ and the batch size of $8$. OD-Net and Rec-Net are separately trained using the mixed binary mask dataset (Sec.~\ref{sssec:od_net_dataset}) and 2D-3D dataset (Sec.~\ref{sssec:deformation_model_dataset}) respectively. In total, we train Rec-Net and OD-Net for $130$ and $30$ epochs respectively. The architecture is implemented using \textit{PyTorch} 
\cite{paszke2017automatic, pytorch2018}. 
In the 2D-3D dataset, we extract $20$ sequential states out of every $100$ consecutive states for testing and use the remaining data for Rec-Net training. Likewise, we divide the binary mask dataset in the ratios ${4}{:}{1}$ for the training and testing of OD-Net. We use mean squared error (MSE) to penalise the discrepancy between the output and the ground truth binary images. %

\begin{table*}[t!]
\parbox{.5\linewidth}{
\centering
\scalebox{0.63}{%
\renewcommand{\arraystretch}{1.3}
 \begin{tabular}{ccccccc} \toprule 
				& Yu \textit{et al.}~\cite{Yu2015} & Liu-Yin \textit{et al.}~\cite{liu2017better} 	& AMP \cite{Golyanik2017d}	& VA \cite{Garg2013} 	& HDM-Net \cite{golyanik2018hdm} & IsMo-GAN \\ \midrule %
      $t$, \textit{sec.}  		& 3.305		& 5.42			& 0.035				& 0.39  		& 0.005	 & \textbf{0.004}	\\ 	
      $e_{3D}$  		& 1.3258		& 1.0049 		& 1.6189			& 0.46			& 0.0251 & \textbf{0.0175}	\\ 
      $\sigma$			&  \textbf{0.007}	&0.0176			& 1.23				& 0.0334		& 0.03 & 0.01		\\ \hline
 \end{tabular}}
 \caption{Reconstruction times per frame $t$ in \textit{seconds}, $e_{3D}$ and standard deviation $\sigma$ for Yu \textit{et al.}~\cite{Yu2015}, Liu-Yin \textit{et al.}~\cite{liu2017better}, AMP \cite{Golyanik2017d}, VA \cite{Garg2013}, HDM-Net \cite{golyanik2018hdm} and our IsMo-GAN method, for the test interval of $400$ frames. } \label{tab:runtimes} 
 \vspace{3pt}
\scalebox{0.80}{%
 \begin{tabular}{ccccccc}\toprule 
		&  & \textit{illum.~1} 	& \textit{illum.~2} 	& \textit{illum.~3} 	& \textit{illum.~4}	& \textit{illum.~5} 	\\ \midrule 
   HDM-Net \cite{golyanik2018hdm}  & $e_{3D}$  & 0.07952	 	& 0.0801		& 0.07942		& 0.07845	& 0.07827\\ 
     				 & $\sigma$	& 0.0525	&  0.0742		& 0.0888		& 0.1009		& 0.1123		\\ \midrule 
    IsMo-GAN 				 & $e_{3D}$  & \textbf{0.06803}	 	& \textbf{0.06908}		& \textbf{0.06737}		& \textbf{0.06754}	& \textbf{0.06685}	\\ 
    					 & $\sigma$	& \textbf{0.0499}	&  \textbf{0.0696}		& \textbf{0.0824}		& \textbf{0.093}		& \textbf{0.102}		\\ \hline       
 \end{tabular}}
\caption{Comparison of 3D error for different illuminations. The \textit{illuminations} $1$-$4$ are known, and the \textit{illumination} $5$ is unknown.  }\label{tab:avg_errors_illuminations}
}
\centering
\hfill
\parbox{.45\linewidth}{
        \centering
	\vspace{-7pt}
  \scalebox{0.83}{
  \renewcommand{\arraystretch}{0.8} %
 \begin{tabular}{cccccc}\toprule 
		&  &\textit{endoscopy} 	& \textit{graffiti} 	& \textit{clothes} 	& \textit{carpet} 	\\ \midrule 
   HDM-Net \cite{golyanik2018hdm}  & $e_{3D}$	&0.0485	& 0.0499		& 0.0489		& 0.1442		\\    
     			     &$\sigma$	&\textbf{0.0135}	&  0.022		& 0.0264		&  0.0269		\\ \midrule 
    IsMo-GAN 				 & $e_{3D}$  & \textbf{0.0336}	 	& \textbf{0.0333} 	& \textbf{0.0353}		& \textbf{0.1105}\\ 
    					 & $\sigma$	& 0.0148	&  \textbf{0.0208}		& \textbf{0.0242	}	& \textbf{0.0268}	\\ \hline 
\end{tabular}}
 \caption{$e_{3D}$ comparison for differently textured surfaces under the same illumination (\textit{illumination} $1$). } 
  \label{tab:avg_errors_textures}
  \vspace{11pt}
  \scalebox{0.68}{%
\renewcommand{\arraystretch}{1.3}
 \begin{tabular}{ccccc}\toprule 
			&Liu-Yin \textit{et al.}~\cite{liu2017better} & Tien Ngo \textit{et al.}~\cite{Ngo_2015_ICCV} & HDM-Net \cite{golyanik2018hdm} & IsMo-GAN 	\\ \midrule 
	$e_{3D}$	& 0.9109  		& 0.0945		& 0.0994		& \textbf{0.0677}	\\    
	$\sigma$& \textbf{0.0677}	&0.1170 	& 0.0809		& \textbf{0.0697	}	\\ \hline 
 \end{tabular}}
 \caption{$e_{3D}$ comparison of the template-based approaches \cite{liu2017better, Ngo_2015_ICCV}, HDM-Net \cite{golyanik2018hdm} 
 and IsMo-GAN on the textureless surfaces from the dataset of Golyanik \textit{et al.}~\cite{golyanik2018hdm}. } 
  \label{tab:avg_errors_Nontextures} 
}
\end{table*}

\section{Experimental Evaluation} 
\label{sec:Experiment}

We evaluate the reconstruction accuracy of IsMo-GAN with different illuminations, textures and occlusions in the input images. 
Our system for training and experiments includes $256$ GB RAM, Intel Xeon CPU E5-2687W v3 running at $3.10$ GHz and GeForce GTX 1080Ti GPU with $11$ GB RAM running under Ubuntu 16.04. %
We compare our framework with three template-based reconstruction methods of Yu \textit{et al.}~\cite{Yu2015}, Liu-Yin \textit{et al.}~\cite{liu2017better} and  
Tien Ngo \textit{et al.}~\cite{Ngo_2015_ICCV}, two NRSfM approaches based on different principles, \textit{i.e.,}  variational NRSfM approach (VA) \cite{Garg2013} and Accelerated Metric Projections (AMP) \cite{Golyanik2017d}, HDM-Net of Golyanik \textit{et al.}~\cite{golyanik2018hdm} and monocular surface reconstruction approach for textureless surfaces of Bedna\v{r}\'{i}k \textit{et al.}~\cite{bednarik2018learning}. \cite{liu2017better} is an extension of \cite{Yu2015} with a shape-from-shading component. For consistency, we adopt the evaluation setting 
as proposed in \cite{golyanik2018hdm} and report the 3D reconstruction error $e_{3D}$ along with the standard deviation of $e_{3D}$ over a set of frames denoted by $\sigma$. $e_{3D}$ is defined as 
\begin{equation}\label{eq:loss_3D} 
e_{3D}=\frac{1}{MN}\sum_{m=1}^{M}\sum_{n=1}^{N}\frac{\lVert\mathbf{S}_{m}^{\text{GT}} - \mathbf{G(I}_m^{n})\rVert_{\mathcal{F}}}{\lVert\mathbf{S}_{m}^{\text{GT}}\rVert_{\mathcal{F}}}, 
\end{equation} 
where $\lVert \cdot \rVert_{\mathcal{F}}$ denotes the Frobenius norm.

\begin{figure}[t]
    \centering
   \includegraphics[width=1.0\linewidth]{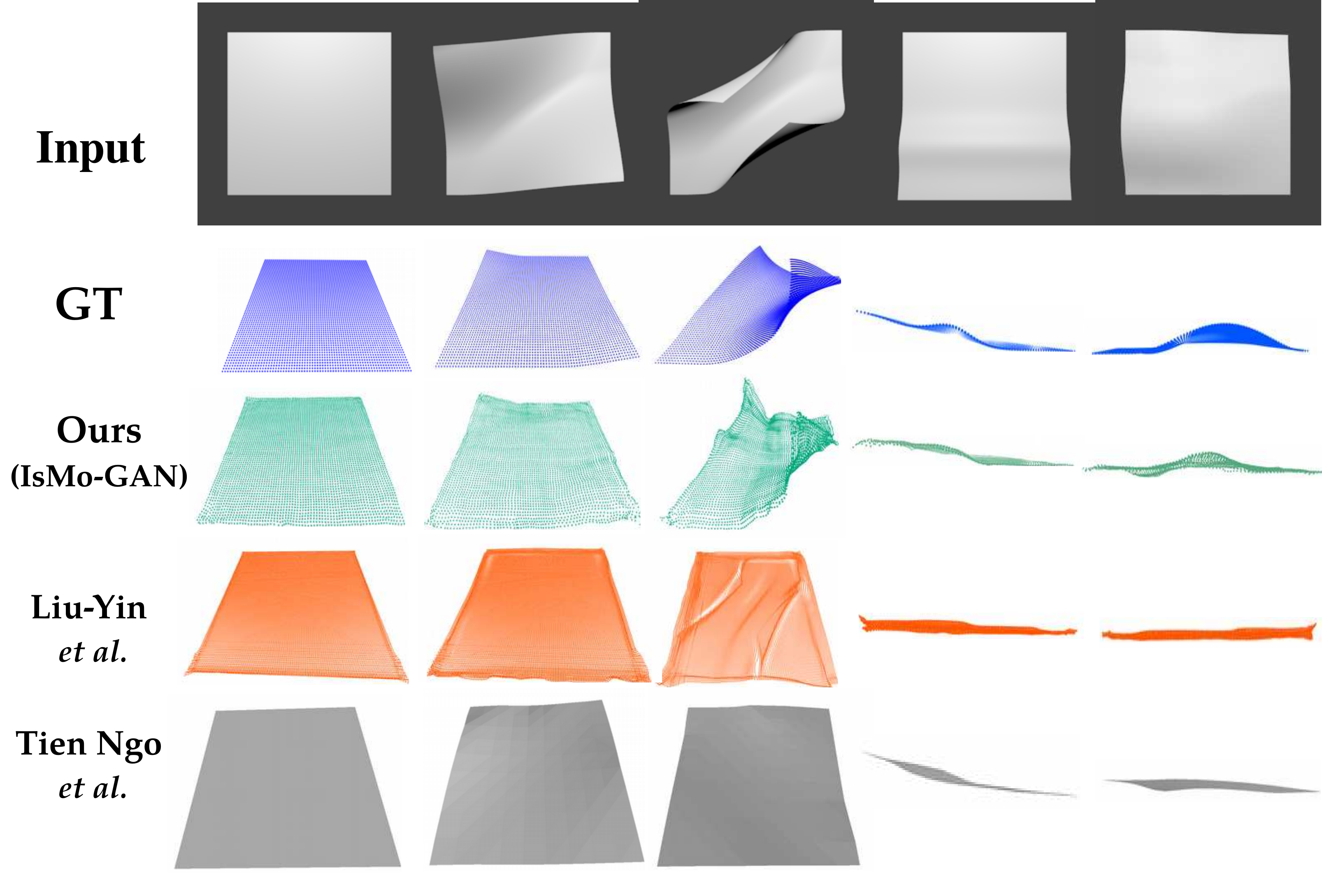} 
    \captionof{figure}{Selected reconstruction results of Liu-Yin \textit{et al.}~\cite{liu2017better}, Tien Ngo \textit{et al.}~\cite{Ngo_2015_ICCV} and IsMo-GAN on the textureless surfaces from the training set. } %
    \label{fig:yucomp}
   \vspace{-7pt}
\end{figure}

\begin{figure*}[t!]
 \centering
  \includegraphics[width=1.0\linewidth]{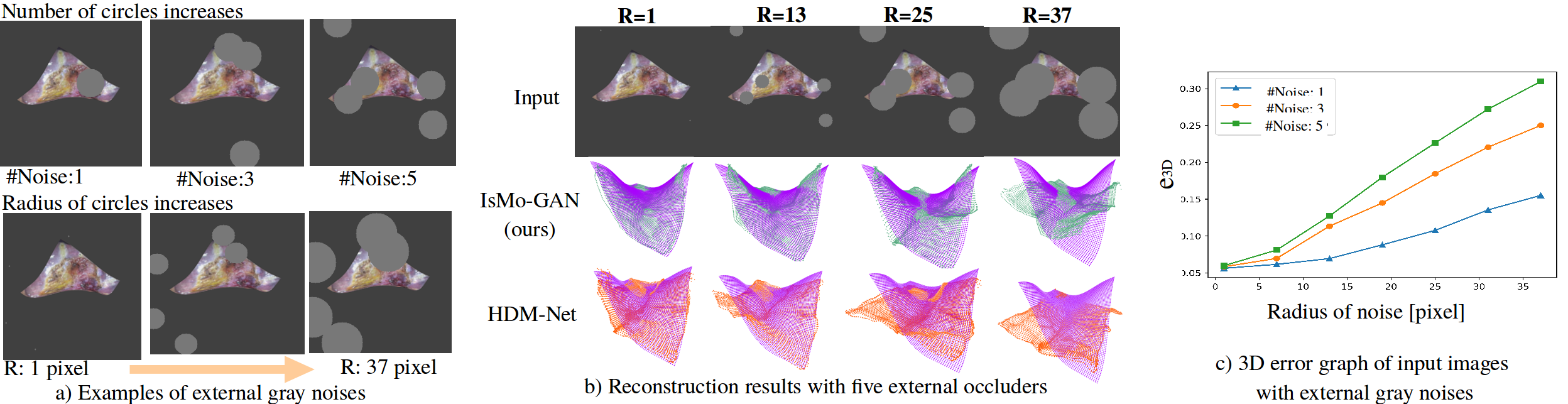} 
  \caption{\textbf{a)} Exemplary occluded images with the increasing number of occluders (the top row) and the increasing size of the occluders (the bottom row). %
  \textbf{b)} Outputs of our network and HDM-Net \cite{golyanik2018hdm} with five external occluders --- 
  ground truth shapes (purple), reconstructions by IsMo-GAN (green) and HDM-Net (orange). %
  \textbf{c)} 3D error graph for images with external occlusions. %
  In \textbf{a)} and \textbf{b)}, $\mathsf{R}$ denotes radii of occluders. Best viewed in colour. 
  } 
  \vspace{8pt}
\label{fig:noseExp_graph} 
\end{figure*} 

\subsection{Synthetic \textit{\large \textbf{Thin Plate}} Dataset \cite{golyanik2018hdm}}\label{ssec:synth_data_experiments} 

Table~\ref{tab:runtimes} summarises the accuracy and the runtimes on a test sub-sequence with $400$ frames chosen 
such that it can be processed by all tested methods. %
AMP \cite{Golyanik2017d} has the highest throughput, and \cite{Garg2013} shows the highest accuracy among non deep learning methods. IsMo-GAN 
outperforms all other methods in the reconstruction accuracy. %
Compared to HDM-Net \cite{golyanik2018hdm}, the runtime improves by $0.001$ seconds per frame on average which means that IsMo-GAN supports processing rates of up to $250$ \textit{Hz} compared to $200$ \textit{Hz} of HDM-Net. 
As shown in Table~\ref{tab:avg_errors_illuminations}, our framework also excels HDM-Net \cite{golyanik2018hdm} in the test with varying illuminations. We do not observe a large difference in $e_{3D}$ for different 
positions of the light source, 
which suggests the enhanced property of illumination invariance. We report $e_{3D}$ for known (\textit{endoscopy}, \textit{graffiti} and \textit{clothes}) and unknown (\textit{carpet}) textures in Table~\ref{tab:avg_errors_textures}. In all cases, our approach outperforms HDM-Net \cite{golyanik2018hdm} reducing the error by $>20\%$ on average. As expected, $e_{3D}$ is higher for the unknown texture compared to the known ones. Still, 
we do not find severe qualitative faults in the reconstructions. 
In the textureless case, our approach shows much lower $e_{3D}$ than Liu-Yin \textit{et al.}~\cite{liu2017better} and $\approx 30\%$ lower $e_{3D}$ than HDM-Net, see Table \ref{tab:avg_errors_Nontextures} and Fig.~\ref{fig:yucomp} with visualisations. Liu-Yin \textit{et al.}~\cite{liu2017better} assume the contour of the target object to be consistent since it uses masking to distinguish the region of interest from the background. Therefore, for a fair comparison, we choose predominantly small deformations from our dataset (see Fig.~\ref{fig:yucomp}). %
Tien Ngo \textit{et al.}~\cite{Ngo_2015_ICCV} support poorly textured surfaces when the observed deformations are rather small. %
All in all, this is a significant improvement compared to the baseline HDM-Net approach \cite{golyanik2018hdm}, as IsMo-GAN uses the same training dataset for the geometry regression as HDM-Net, while relying on other regression criteria (\textit{e.g.,} adversarial loss).

\paragraph{External Occlusions. } Next, we evaluate IsMo-GAN in the scenario with external occlusions. %
We select an arbitrary 3D state from the test dataset with a comparably large deformation and introduce random circular noise (grey  circles) into the corresponding 2D images. The size and the number of occluders vary as shown in Fig.~\ref{fig:noseExp_graph}-(a). %
We show the reconstruction results with five introduced occluders in Fig.~\ref{fig:noseExp_graph}-(b).
For each combination of the occluder's size and the number of occluders, we generate ten images and report 
the average $e_{3D}$ of the IsMo-GAN reconstructions for these images, see Fig.~\ref{fig:noseExp_graph}-(c). 
Unless the input image contains large occlusions, our network keeps the high reconstruction accuracy. When the occluder's size reaches $7$ pixels, the slope of the graph increases which marks the robustness threshold, with up to $40\%$ of the object being occluded. 

\subsection{Real \textit{\large \textbf{Textureless Cloth}} Dataset \cite{bednarik2018learning}}\label{subs:textureless} 

\begin{figure}[t!] 
    \centering
    \includegraphics[width=1.0\linewidth]{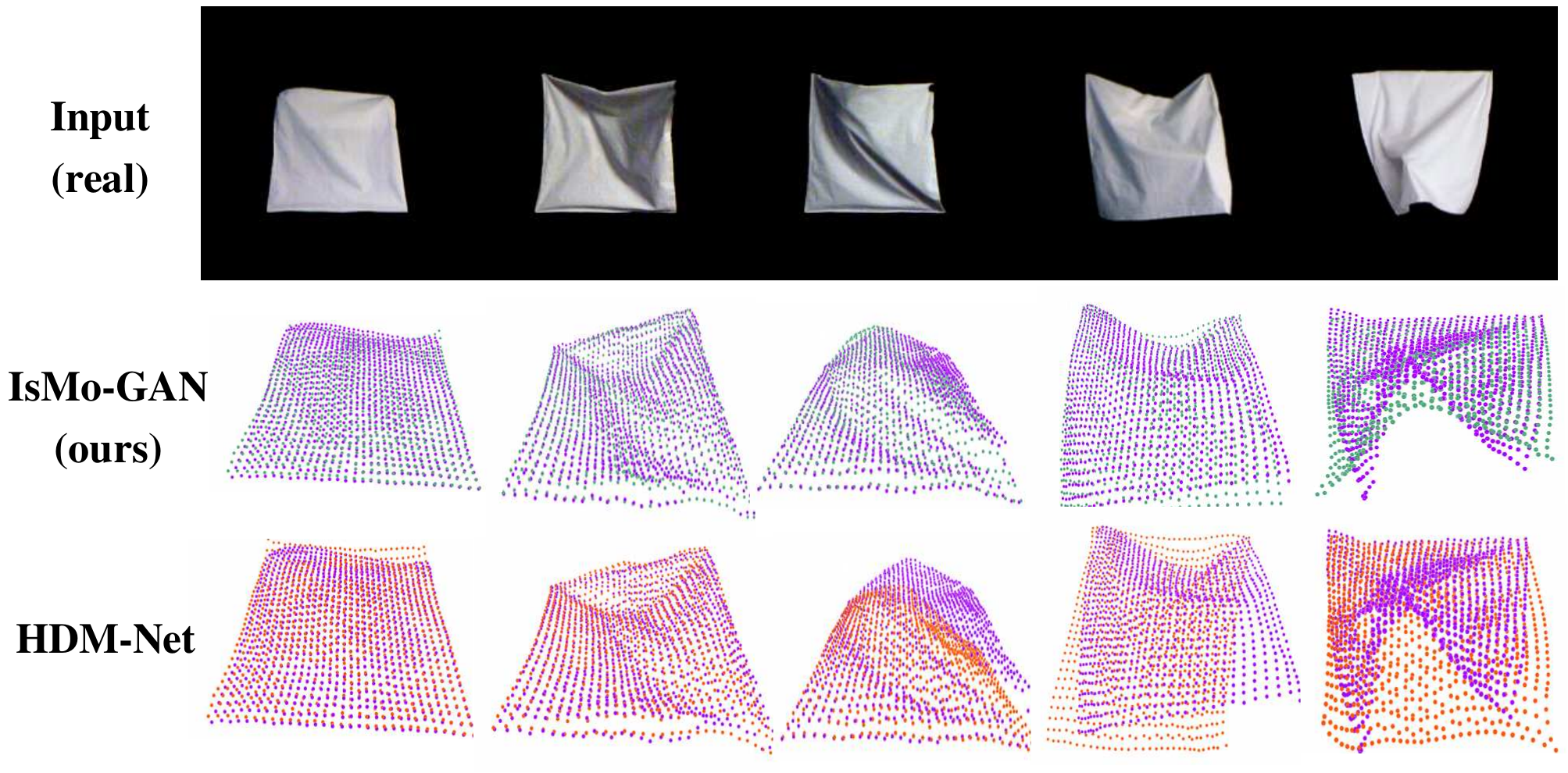} %
    \captionof{figure}{Selected reconstructions of the textureless \textit{cloth} dataset \cite{bednarik2018learning}.} 
    \label{fig:textureless_dataset}
\end{figure}

We also evaluate IsMo-GAN on the real \textit{cloth} dataset \cite{bednarik2018learning} with textureless deforming surfaces with varying shading. 
For every frame, the dataset includes ground truth meshes of the observed surfaces (with $31^2$ points per state) obtained by fitting 
a mesh template to the captured depth maps \cite{bednarik2018learning}. Similarly to the evaluation with the thin plate dataset \cite{golyanik2018hdm}, 
we split all frames in the propotion $80$-$20\%$ for the training and test subsets respectively. %
Since the \textit{cloth} dataset contains $6237$ samples and is smaller than the thin plate dataset, we omit two layer blocks in the generator's encoder (sets of convolutions, batch normalisation, leaky ReLU and max pooling) as well as two layer blocks in the generator's decoder (sets with deconvolutions, batch normalisation and leaky ReLU) and adjust the kernel sizes. The dimensionality of the latent space is reduced to $11\times11\times256$. 

We compare the proposed IsMo-GAN with HDM-Net \cite{golyanik2018hdm} and the monocular 3D reconstruction approach for non-rigid textureless surfaces 
of Bedna\v{r}\'{i}k \textit{et al.}~\cite{bednarik2018learning}. While Bedna\v{r}\'{i}k \textit{et al.}~report the SAD of $21.48$ $mm$ \cite{bednarik2018learning}, %
HDM-Net \cite{golyanik2018hdm}~achieves $17.65$ $mm$. SAD of our IsMo-GAN amounts to $15.79$ $mm$ which is a $26.5\%$ improvement in comparison to 
Bedna\v{r}\'{i}k \textit{et al.}~\cite{bednarik2018learning}\footnote{note that details on the dataset split are not provided in \cite{bednarik2018learning}}, and a $10.5\%$ improvement versus HDM-Net \cite{golyanik2018hdm}. %
Compared to Bedna\v{r}\'{i}k \textit{et al.}~\cite{bednarik2018learning}, we use deconvolutional layers in the decoder instead of the fully-connected 
layers. We believe that point adjacencies provide a strong cue for surface reconstruction. %
Fig.~\ref{fig:textureless_dataset} shows selected reconstructions of challenging states. Even though SAD of HDM-Net is just $1.86$ $mm$ larger 
as compared to IsMo-GAN on average, HDM-Net often fails to reconstruct states with large folds and deformations. %
Our architecture is not restricted to globally smooth surfaces and captures fine geometric details revealed by the shading cue.

\subsection{Real Images (Qualitative Results)}\label{sec:real_images} %

\begin{figure}[t!]
    \centering
    \includegraphics[width=1.0\linewidth]{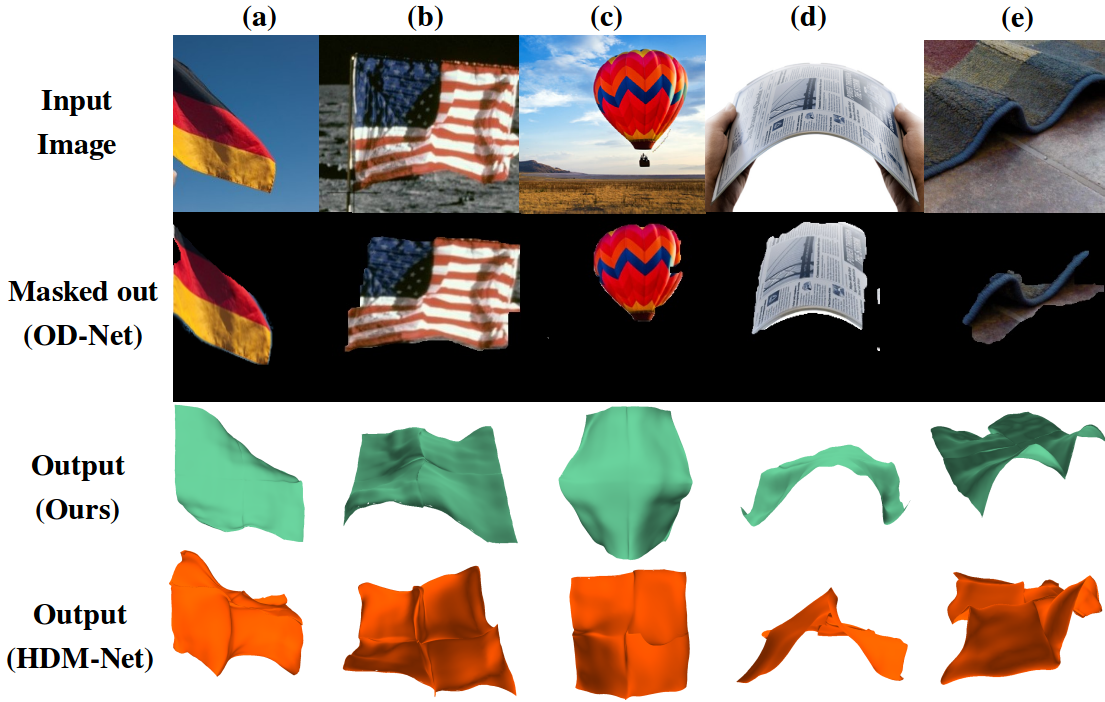}
    \captionof{figure}{3D reconstruction results from real images: a German flag \cite{germanflag}, an American flag \cite{americanflag}, a hot air balloon \cite{balloon}, a bent surface \cite{eink} and a carpet with a double wrinkle \cite{wrinkle}. All input images are unknown to our pipeline. Note the qualitative improvement in the results of IsMo-GAN compared to the previous HDM-Net method \cite{golyanik2018hdm}. Best viewed enlarged. } %
    \label{fig:real_world} 
\end{figure}

Next, we evaluate IsMo-GAN on a collection of real images. 
In comparison to HDM-Net \cite{golyanik2018hdm}, the strength of IsMo-GAN is the enhanced generalisability to real data, even though 
the deformation model is trained on the synthetic dataset. Fig.~\ref{fig:real_world} 
shows several reconstructions from real images by HDM-Net \cite{golyanik2018hdm} and IsMo-GAN. 
We choose images with a different textures, deformations, illuminations and scene context, \textit{i.e.,} waving flags, a hot air balloon, a bent paper, and a carpet with wrinkles. %
The reconstructions capture well the variety of exemplary shapes. 
None of the textures (Fig.~\ref{fig:real_world}-(a),(b)) 
were present in the training dataset, and IsMo-GAN captures well the main deformation mode and shape. 
The scene with the hot air balloon (Fig.~\ref{fig:real_world}-(c)) has an inhomogeneous background. Thanks to the OD-Net, IsMo-GAN generates qualitatively a more realistic reconstruction than HDM-Net.  Fig.~\ref{fig:real_world}-(e) is an example of a deformation state which is the most dissimilar to the states in the training dataset. Remarkably, our approach recovers the rough geometry of the object in the scene whereas HDM-Net fails to capture it. 

\begin{figure}[t!]
    \centering
    \includegraphics[width=1.0\linewidth]{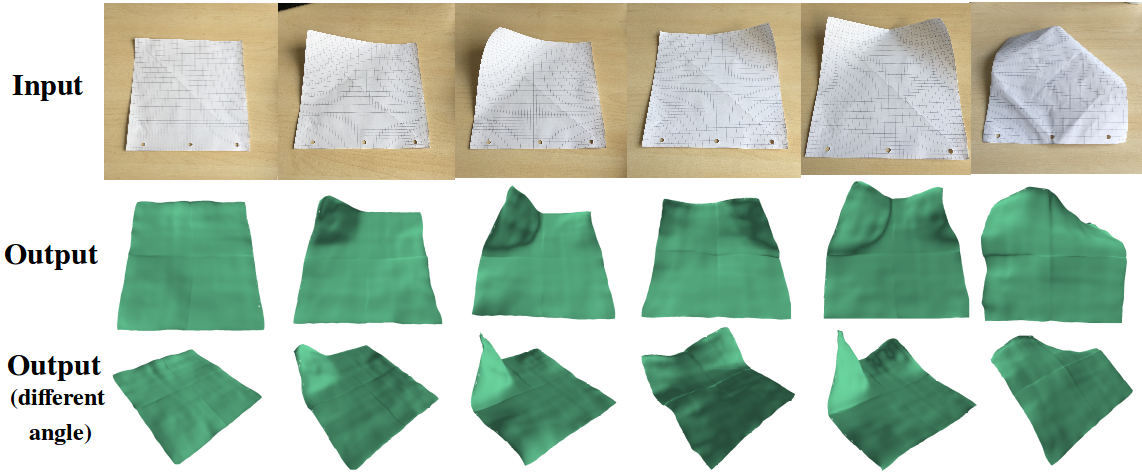}
    \captionof{figure}{3D reconstruction results by IsMo-GAN on the new real \textit{origami} video sequence. 
    Best viewed enlarged. } %
    \label{fig:origami_seq}
\end{figure}

Fig.~\ref{fig:origami_seq} shows the reconstruction results by IsMo-GAN on the new \textit{origami} video sequence. For \textit{origami}, the main reconstruction cue is shading. 
Our approach captures well the global deformation of the target object with a weak texture in the real-world scene. %
Even though IsMo-GAN operates on individual images, the resulting dynamic reconstruction is temporally smooth.


\section{Discussion} %
\label{sec:Limitations} 

The experiments demonstrate the significant qualitative improvement of IsMo-GAN when reconstructing from real images compared to the previous most related method HDM-Net \cite{golyanik2018hdm}. 
We can reconstruct surfaces more accurately in the challenging cases with external occlusions and lack of texture. 
The experiment with textureless \textit{cloth} dataset \cite{bednarik2018learning} in Sec.~\ref{subs:textureless} shows that our pipeline generalises well, can be easily adjusted for other scenarios (\textit{e.g.,} different primary reconstruction cues, surface properties, types of deformations, \textit{etc.}) 
and even outperform competing specialised methods. Even though we do not explicitly assume gradual frame-to-frame surface deformations, IsMo-GAN recovers temporally smooth surfaces from a video sequence as shown in Sec.~\ref{sec:real_images}. 
Especially the enhanced accuracy for textureless surfaces is a valuable property in passive 3D capture devices operating in real human-made environments. The inference in IsMo-GAN is light-weight (running at $250$ $Hz$) and would require low energy, making it appealing for mobile augmented reality devices. 

IsMo-GAN shows plausible results especially when similar non-rigid states appear in the training dataset or when the target state can be represented as a 
blend of known deformation states. Otherwise, IsMo-GAN can be retrained with a dataset encoding another deformation model or covering more deformation modes, as has been demonstrated in Sec.~\ref{subs:textureless}. 
Moreover, the accuracy of our approach depends on the accuracy of the binary mask generation in the real-world scenario, and this aspect can also be improved for pre-defined scenarios.

\section{Conclusion} 
\label{sec:Conclusion} 

In this study, we introduce IsMo-GAN --- the first DNN-based framework for deformation model-aware non-rigid 3D surface regression from single monocular images with point set representation trained in an adversarial manner. %
The proposed approach regresses realistic general non-rigid surfaces from real images while being trained on a synthetic dataset of non-rigid states with varying light sources, textures and camera poses. 
Compared to the previously proposed DNN based methods \cite{golyanik2018hdm, pumarola2018geometry}, our pipeline localises the target object with an OD-Net. 
Thanks to the point cloud representation, we take advantage of computationally efficient 2D convolutions. 

In the extensive experiments, IsMo-GAN outperforms competing 
methods, both model-based and DNN-based, in the reconstruction accuracy, throughput, robustness to occlusions as well as the ability to handle textureless surfaces. 
In future work, we plan to collect more real data and test IsMo-GAN in the context of medical applications. %
For video sequences such as \textit{origami} reconstructed in Sec.~\ref{sec:real_images}, a temporal smoothness term could further 
improve the results. 
Another future direction is network pruning for deployment of IsMo-GAN on an embedded device. 
Besides, a superordinate system can include IsMo-GAN as a component for shape recognition or surface augmentation. %

\section*{Acknowledgement} 
This work was supported by project VIDETE (01IW18002) of the the German Federal Ministry of Education and Research (BMBF) and  the ERC Consolidator Grant 4DReply (770784) .

\appendix 

{\small
\bibliographystyle{ieee}
\bibliography{egbib}
}
\begin{figure*}[t!]
    \centering
    \includegraphics[width=1.01\linewidth]{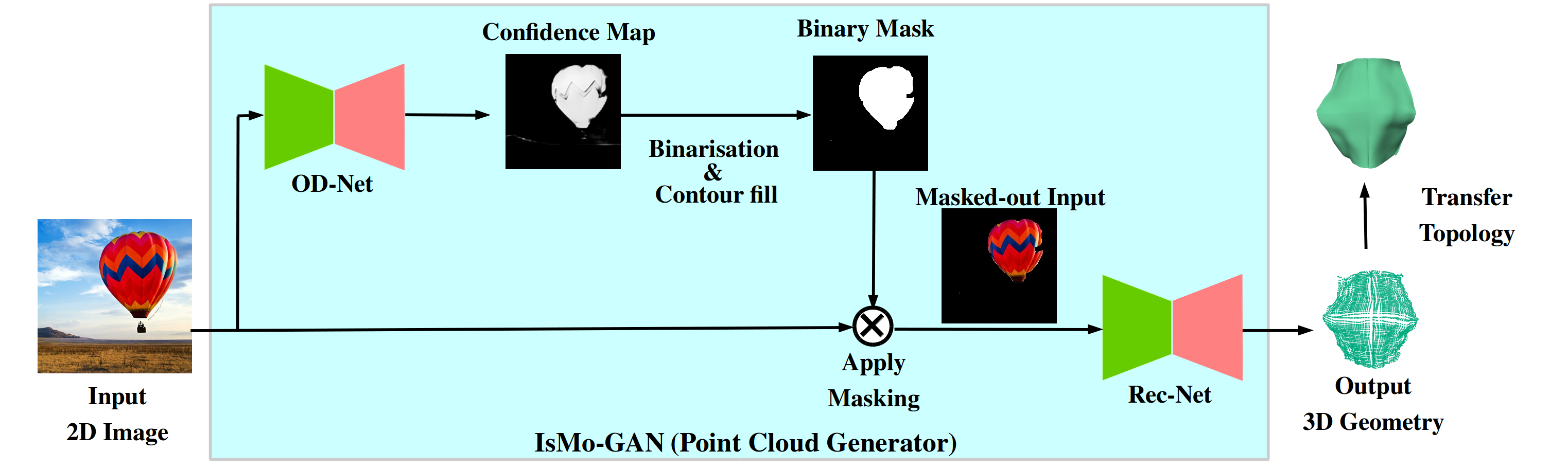}
    \captionof{figure}{A high-level overview of the proposed IsMo-GAN reconstruction pipeline. OD-Net accepts an input RGB image and generates an initial object-background confidence map. The confidence map is further converted into a binary segmentation mask. Rec-Net accepts a segmented RGB image as an input and infers 3D surface geometry. 
    Since the topology is consistent throughout all states, it can be transferred to the output. 
} %
\vspace{12pt} 
\label{fig:overview}
\end{figure*}
\begin{figure*}[t!]\begin{minipage}{.5\linewidth}
\centering
\subfloat[]{\label{main:a}\includegraphics[width=\linewidth]{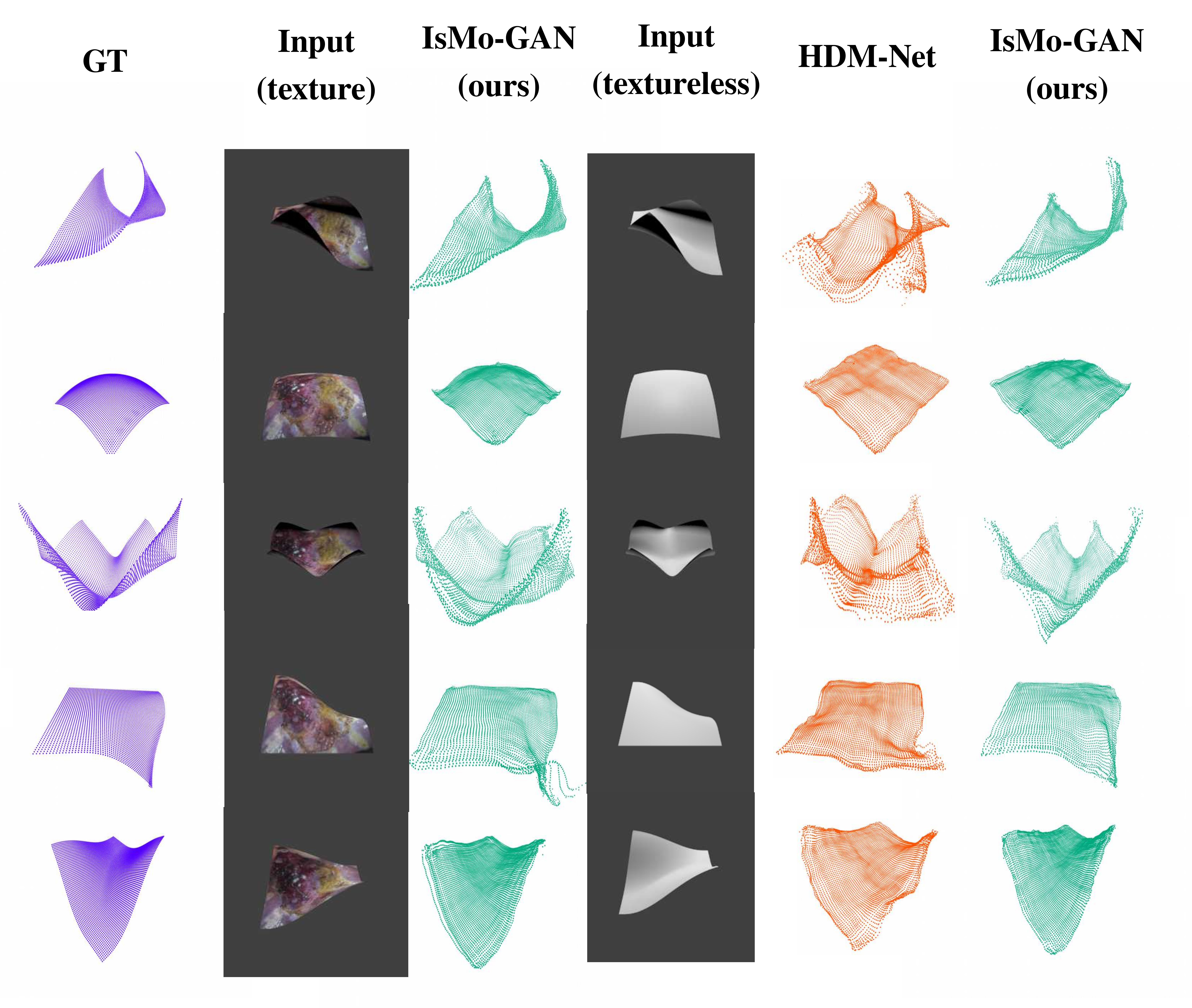}}
\end{minipage}%
\begin{minipage}{.5\linewidth}
\centering
\subfloat[]{\label{main:b}\includegraphics[width=\linewidth]{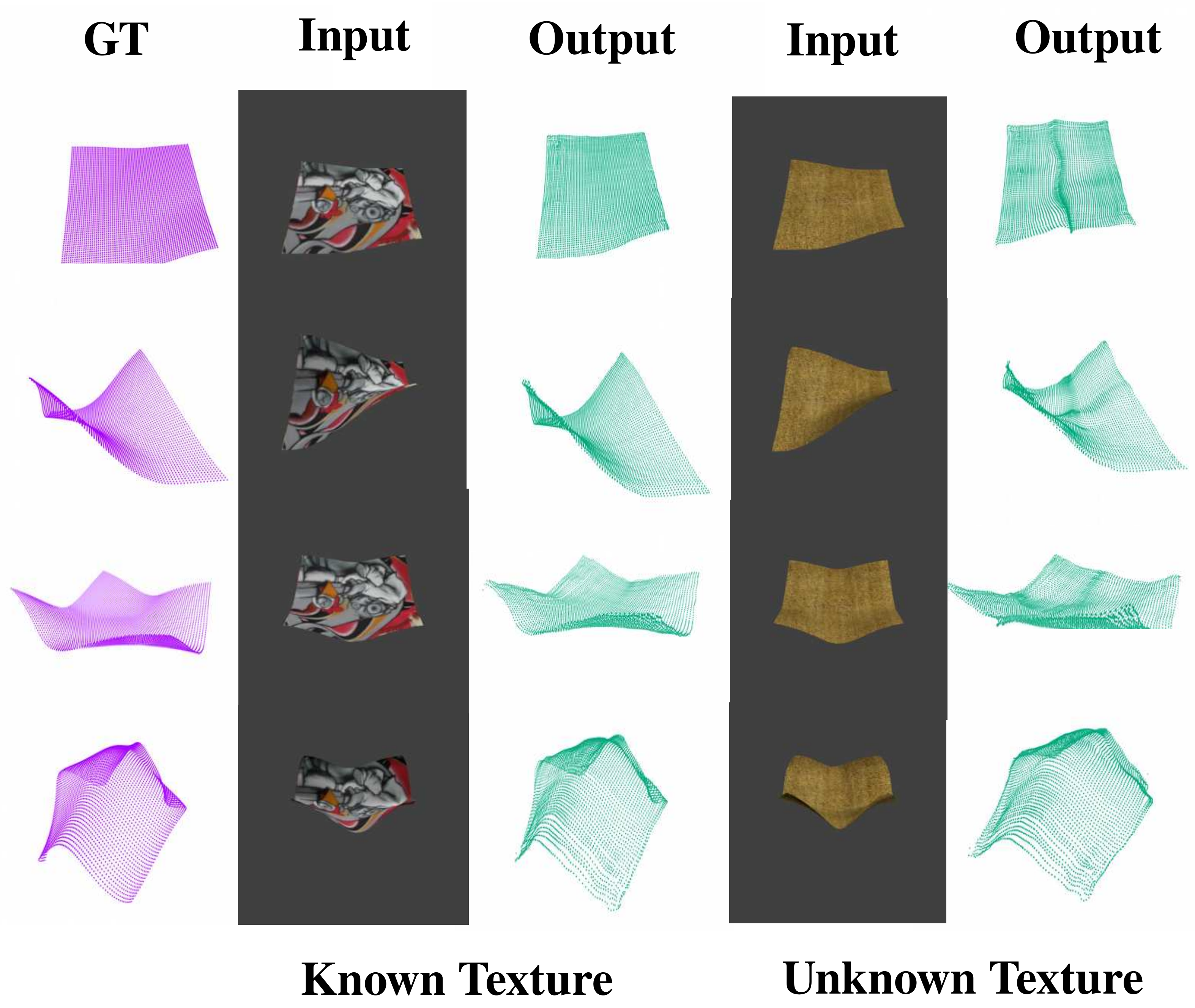}}
\end{minipage}
\caption{\textbf{(a)} Selected reconstruction results of our framework on endoscopically textured and textureless surfaces for unknown states from the training set. The textured input reconstructions of HDM-Net are not shown due to no significant differences from the ones of IsMo-GAN. \textbf{(b)} 3D reconstructions by our method with known and unknown texture inputs. GT stands for ground truth. \vspace{7pt}}
\label{fig:comparison_nonTex_Tex_comparison}
\end{figure*}

\section{Supplementary Material} 

In this supplementary material, we provide a schematic high-level overview 
of the proposed IsMo-GAN framework in Sec.~\ref{app:scheme} and show more results on the \textit{thin plate} dataset \cite{golyanik2018hdm} in  Sec.~\ref{app:thin_plate} and the \textit{textureless cloth} dataset \cite{bednarik2018learning} in Sec.~\ref{app:textureless_cloth}. 
We use the same list of references as in the main matter (please see the \textit{References} section above). 

\subsection{A High-Level Visualisation}\label{app:scheme} %

Fig.~\ref{fig:overview} provides a high-level visualisation of the entire IsMo-GAN pipeline  
including OD-Net for the confidence map estimation (the background-foreground segmentation) and Rec-Net. 
The initially probabilistic estimates of OD-Net are firstly binarised by the algorithm of \cite{4310076}, and then all pixels inside the same contour are filled with white (indicating the object) using the border following algorithm \cite{suzuki1985topological}. 
Finally, the binary mask is applied to the input 2D RGB image to extract the target region, and the segmented image is passed to Rec-Net. 
Since the topology of the object at rest is known in the training dataset \cite{golyanik2018hdm} and does not change while the surface is deforming, we transfer it to the output point set. 

\begin{figure*}[th] %
    \centering
    \includegraphics[width=1.0\linewidth]{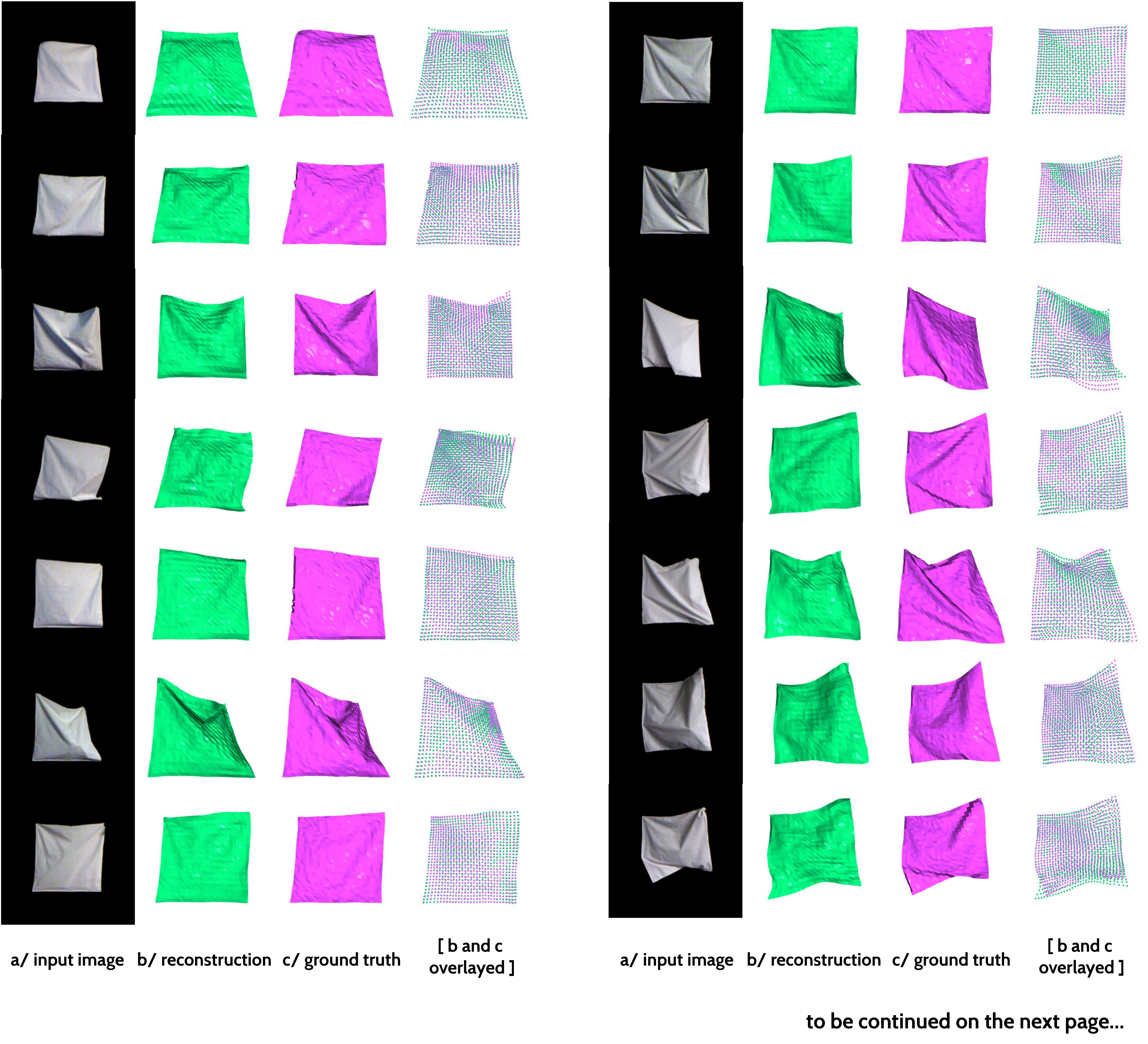}
\end{figure*}

\subsection{Further Comparisons on the \textit{\large \textbf{Thin Plate}} \cite{golyanik2018hdm}}\label{app:thin_plate} %

Selected reconstructions of IsMo-GAN on endoscopically textured and textureless surfaces from the \textit{thin plate} dataset \cite{golyanik2018hdm} 
are shown in Fig.~\ref{fig:comparison_nonTex_Tex_comparison}. 
Our approach reconstructs surfaces of high visual accuracy in the textureless case (see Fig.~\ref{fig:comparison_nonTex_Tex_comparison}-(a)), whereas HDM-Net
often distorts the structure too much. %
Note that we retrain HDM-Net with the textureless surfaces (our dataset extension) for a fair comparison. 

Next, as expected, the IsMo-GAN's error for an unknown texture (\textit{carpet}) is slightly higher compared to the known textures (\textit{cf.}~Table~\ref{tab:avg_errors_textures}). 
Still, the overall states are correctly recovered, and the surfaces are smooth in most of the cases, thanks to the adversarial regulariser (see Fig.~\ref{fig:comparison_nonTex_Tex_comparison}-(b)). 

\subsection{More Results on the \textit{\large \textbf{Textureless Cloth}} \cite{bednarik2018learning}}\label{app:textureless_cloth} %
\begin{figure*}[t!]
    \centering
    \includegraphics[width=1.0\linewidth]{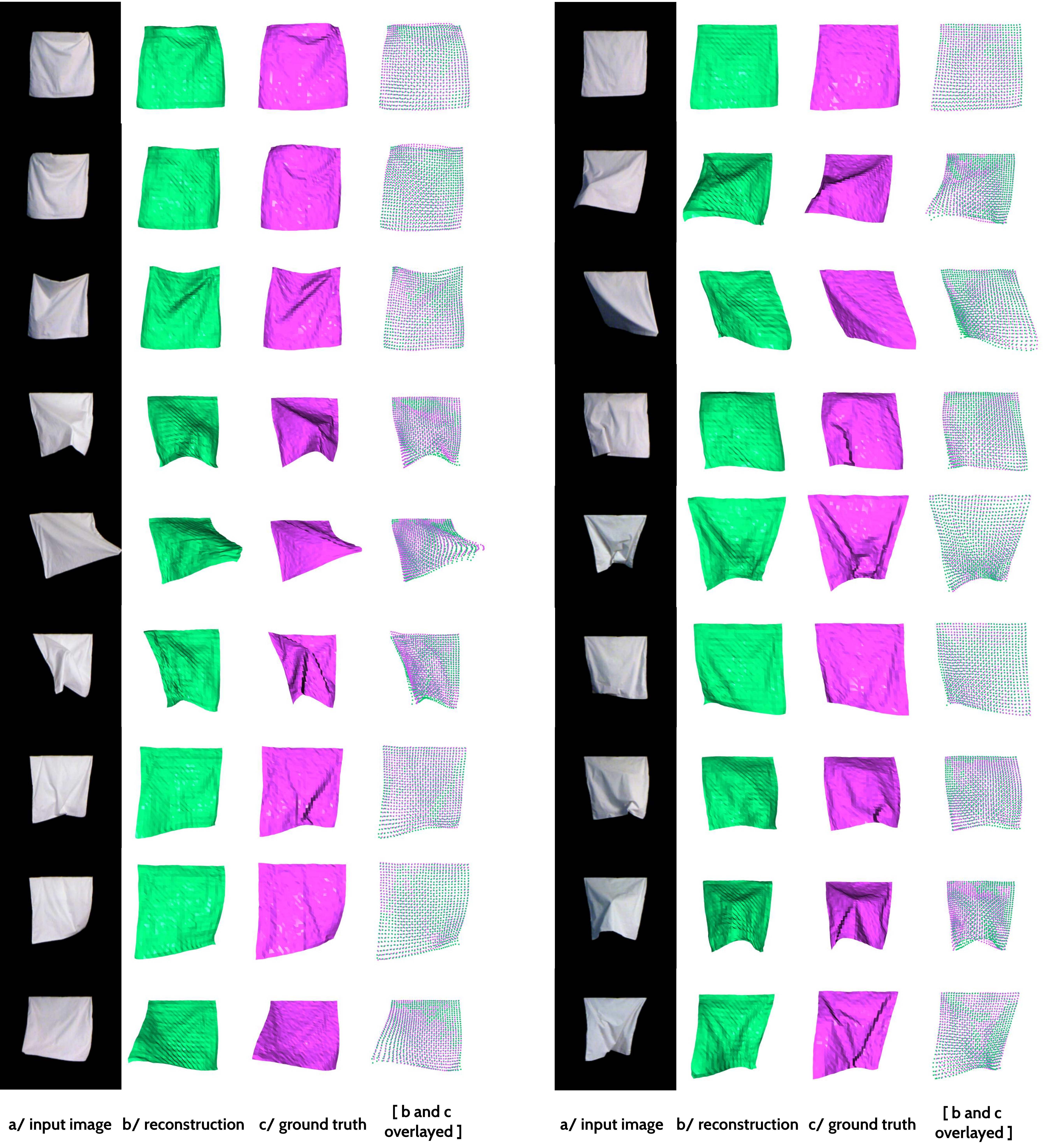}
    \vspace{5pt}
    \captionof{figure}{Selected reconstructions of our IsMo-GAN approach on the real \textit{textureless cloth} dataset \cite{bednarik2018learning}, with 
    the average SAD of $15.79$ $mm$. 
    \textbf{(a)}: Input RGB images, \textbf{(b)}: output of IsMo-GAN, \textbf{(c)}: ground truth, \textbf{the right-most column}: overlay of our reconstructions and the ground truth. } 
    \label{fig:3dvcomp} 
\end{figure*}

We show $32$ additional reconstructions by IsMo-GAN on real \textit{textureless cloth} dataset \cite{bednarik2018learning} in Fig.~\ref{fig:3dvcomp} 
(our estimates are in \textit{green}, and the ground truth is in \textit{magenta}). 
Our approach accurately captures the overall cloth deformations, and large wrinkles are reconstructed in most of the cases. 
Note that compared to the \textit{thin plate} dataset \cite{golyanik2018hdm}, the states are changing faster and discretely in the \textit{textureless cloth} dataset \cite{bednarik2018learning} (the states were created by manually deforming the cloth). 
Although we keep the training-test split as in the case of the \textit{thin plate} (${4}{:}{1}$), the generalisation of IsMo-GAN is not impeded, and, as shown in Sec.~\ref{subs:textureless}, we outperform both HDM-Net \cite{golyanik2018hdm} and the approach of Bedna\v{r}\'{i}k \textit{et al.}~\cite{bednarik2018learning}. %

\end{document}